\documentclass[runningheads]{llncs}
\usepackage{graphicx}
\usepackage{graphicx}
\usepackage{multirow}
\usepackage{booktabs}
\usepackage{bm}
\usepackage[colorlinks,
            linkcolor=black,
            anchorcolor=black,
            citecolor=black,
            urlcolor=black,
            ]{hyperref}

\begin{document}
\title{A Method of Query Graph Reranking for Knowledge Base Question Answering}
\titlerunning{A Method of Query Graph Reranking for KBQA}
%

\author{Yonghui Jia, Wenliang Chen}
%
%
\institute{School of Computer Science and Technology, Soochow University, China
\email{yhjia2018@163.com, wlchen@suda.edu.cn}}

%
	

\maketitle              

\begin{abstract}

This paper presents a novel reranking method to better choose the optimal query graph, a sub-graph of knowledge graph, to retrieve the answer for an input question in Knowledge Base Question Answering (KBQA). Existing methods suffer from a severe problem that there is a significant gap between top-1 performance and the oracle score of top-n results. To address this problem, our method divides the choosing procedure into two steps: query graph ranking and query graph reranking. In the first step, we provide top-n query graphs for each question. Then we propose to rerank the top-n query graphs by combining with the information of answer type. Experimental results on two widely used datasets show that our proposed method achieves the best results on the WebQuestions dataset and the second best on the ComplexQuestions dataset.


\keywords{Knowledge base question answering  \and Query graph ranking \and Query graph reranking.}
\end{abstract}
\section{Introduction}
Knowledge Base Question Answering (KBQA) is defined to be a task that answers natural language questions over knowledge bases, such as Freebase~\cite{bollacker2008freebase} and Wikidata~\cite{vrandevcic2014wikidata}.
A representative stream of approaches to KBQA is based on semantic parsing~\cite{hu2018state,reddy2014large,sun2020sparqa} which translates textual questions into some semantic representation. To this end, formal meaning representations such as $\lambda-DCS$~\cite{liang2013lambda} are widely adopted to represent semantics of questions. However, such meaning representations need to be mapped to logical constants or an ontology before they can be used to retrieve answers from the knowledge base. The mapping between the meaning representations and an ontology is restricted by the coverage of the ontology~\cite{kwiatkowski2013scaling}.

An alternative to formal meaning representations is to represent question semantics with query graph~\cite{luo2018knowledge,yih2015semantic,yih2016value}, which can be regarded as a sub-graph of knowledge base. This way, the restriction incurred by the coverage of an ontology is overcome. With query graph as the semantic representation, the process of KBQA can be divided into two steps: query graph generation and query graph selection. During query graph generation, the question is parsed into a set of candidate query graphs~\cite{yih2015semantic}. In query graph selection, an optimal query graph is selected through ranking methods, and the knowledge graph node corresponding to the optimal query graph is returned as the answer to the question. In the process described above, query graph generation determines the upper bound of the performance that we can achieve whereas query graph selection determines the final performance of a KBQA system. In this paper, we put focus on question graph selection which is thought to be more important and challenging.

Existing approaches to query graph selection generally build on single-stage ranking~\cite{bao2016constraint,luo2018knowledge,yih2015semantic}, which sorts query graphs according to the similarity between the question and the query graph. There are two typical methods to calculate the similarity according to whether we decompose the query graph. One method is to decompose the query graph into different sub-paths and then calculates the similarity between the question and each sub-path~\cite{yih2015semantic}. The other is designed to measure the similarity between the question and a global representation of the whole query graph~\cite{luo2018knowledge}. The strategy of single-stage ranking is widely adopted, but there is room left for further improvement of query graph selection. The claim is inspired by the observation from our preliminary experiments that there is a significant gap between top-1 performance and the oracle score of top-n candidates.~\footnote{By {\em oracle score of top-n candidates}, we mean the performance that we can always choose the optimal query graph from the top-n set.} 
In addition, error analysis shows that some bad cases have top-1 query graphs that retrieve answers of erroneous answer types.

To address the above problems, we propose to use a ranking-reranking strategy for query graph selection. 
Specifically, in the ranking stage we convert the query graph into a sequence and use BERT~\cite{devlin2018bert} to conduct semantic matching between the question and the query graph sequence. In the reranking stage, we incorporate more information, including answer types to sort the top-n candidates that are produced in the ranking stage. Extensive experiments on benchmark data demonstrate the effectiveness of the reranking stage and the incorporation of answer types.


Our contributions in the paper are summarized as follows,
\begin{itemize}
    \item[$\bullet$] We propose a ranking-reranking method for query graph selection in the KBQA system. This two-stage strategy can rerank top-n query graphs more effectively and improve top-1 performance.
    \item[$\bullet$] Focusing on the answer type error in query graph ranking, we propose to combine the answer type feature and the query graph sequence feature to rerank top-n query graphs. Experiments demonstrate that the method can alleviate the type error in query graph ranking and make a big improvement.
\end{itemize}

\begin{figure*}
\centering
\includegraphics[width=12cm]{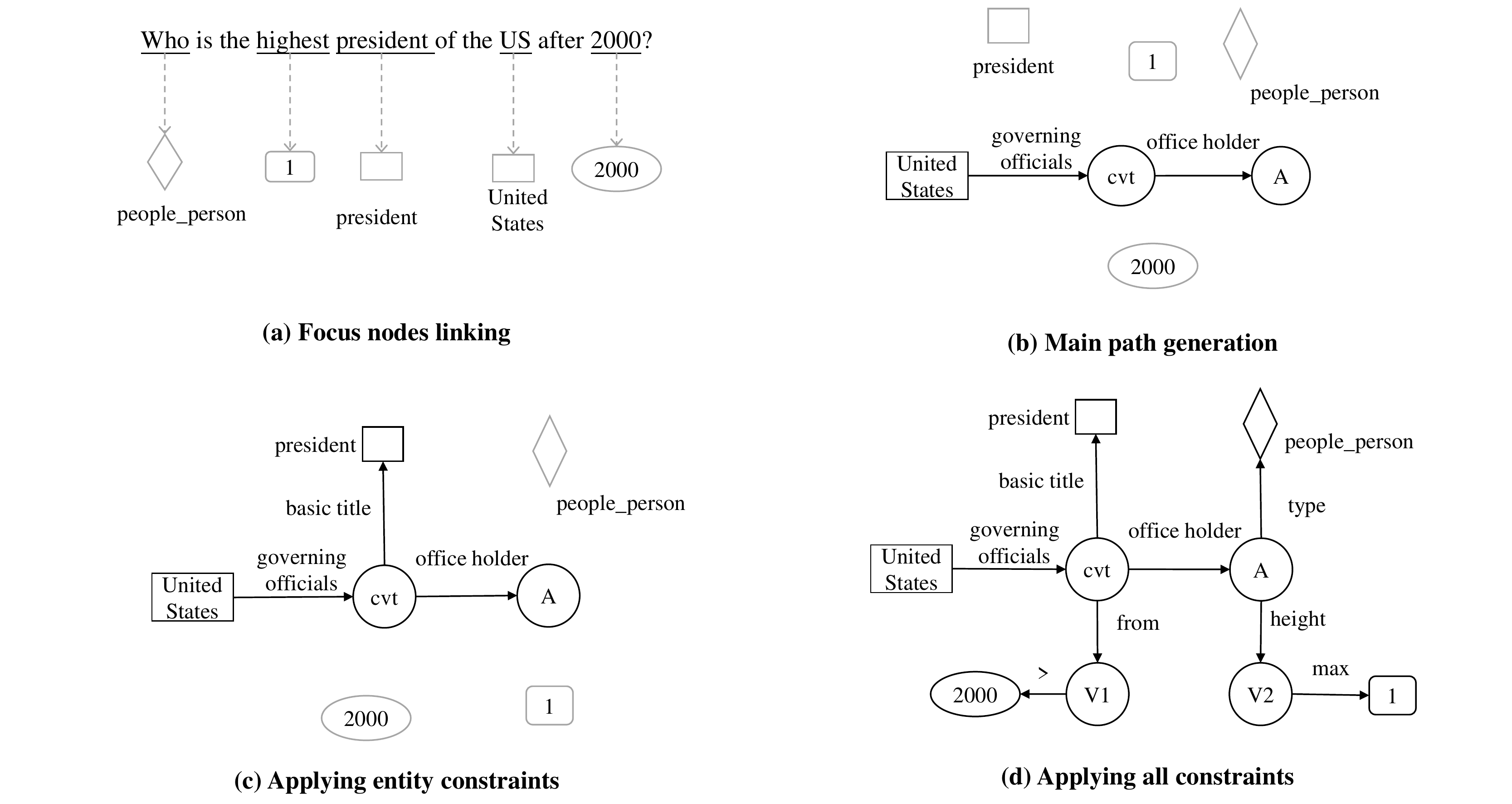}
\caption{Query graph generation procedure of ``Who is the highest president of the US after 2000?''. Figure (a) shows the focus nodes linking result, Figure (b) shows the main path example, Figure (c) shows the result after adding entity constraints to the main path, and Figure (d) shows the query graph after adding all constraints.}
\label{fig:query_graph_generation}
\end{figure*}
\section{Our Approach}
Our approach belongs to the semantic parsing method based on query graph in the field of KBQA. The system includes two main components: query graph generation and query graph selection. In query graph generation, we utilize the staged query graph generation method to obtain a set of query graph candidates. In query graph selection, we propose a two-stage method to select the optimal query graph from the candidates. In the first stage, we rank the candidates to get top-n query graphs, while in the second stage we rerank the top-n query graphs to obtain the best one as the optimal query graph. Finally, the optimal query graph is used to retrieve the final answer from Knowledge Base (KB).


\subsection{Query Graph Generation}
The goal of query graph generation is to parse the question into the form of query graph corresponding to the underlying KB. Here we briefly describe the query graph generation method proposed in the previous studies and the details can be found in the related papers~\cite{luo2018knowledge,yih2015semantic}.

Given a question $q$, we first conduct focus nodes linking to recognize four node constraints, including entities, type words, time words, and ordinal words. For entity linking, we adopt the entity linking tool SMART~\cite{yang2016s} to obtain $<$mention, entity$>$ pairs. For type word linking, we calculate word vector similarity scores between consecutive sub-sequences (up to three words) in the question and all types in KB, and select top-10 $<$mention,type$>$ pairs as type candidates. For time word linking, we use regular expressions to extract time words from the question. For ordinal word linking, we use a predefined ordinal number dictionary and the ``number + superlative" pattern to extract ordinal numbers. Figure 2(a) shows an example of the focus nodes linking.


After focus nodes linking, we perform one-hop and two-hop searches to obtain the main path based on the linked entities. Figure 2(b) shows an example. Then, the entity constraint is added to the main path as a constraint node, as shown in Figure 2(c). After that, we add type constraint, time constraint, and ordinal constraint in turn, and finally get a complete query graph, as shown in Figure 2(d).

During the above procedure, the focus nodes are not fully disambiguated. Moreover, the one-hop and two-hop searches may have different paths. Thus query graph generation may generate more than one, usually hundreds of candidate query graphs. In the circumstances, we can get the candidate query graph set $G=\{g_1,g_2,...,g_n\}$ for each question.

\subsection{Query Graph Selection}

The goal of query graph selection is to select the optimal query graph $g^*$ from the candidate query graph set $G$. In this section, we first describe how to select top-n query graphs from the candidate set by query graph ranking. Then we present a query graph reranking method with answer type to choose the best one from top-n candidates.
\subsubsection{Query Graph Ranking}
In this step, we sort the candidates by calculating the matching score between the question and each candidate query graph, and obtain the top-n query graphs for reranking. To encode the question and a query graph, we choose to convert the query graph into a query graph sequence, which makes the query graph and the question both in the unified sequence form. This choice simplifies the matching strategy of the question and the query graph, and the mature sequence matching methods can be used.


Specifically, w
e first transform the query graph into the sequence form according to its composition. In structure, a query graph consists of up to five sub-paths: MainPath, TypePath, EntityPath, TimePath, and OrdinalPath, corresponding to the main path and four types of node constraints respectively. Each sub-path can be converted into a word sequence. Taking the main path in Figure~\ref{fig:query_graph_generation} as an example, the corresponding main path sequence is ``united states governing officials office holder A". We combine the five sub-path sequences into a query graph sequence according to a fixed sub-path order, denoted as $g^s=\{u_1,u_2,...,u_n\}$, where $u_i$ is a word.

\begin{figure*}
\centering
\includegraphics[width=12cm]{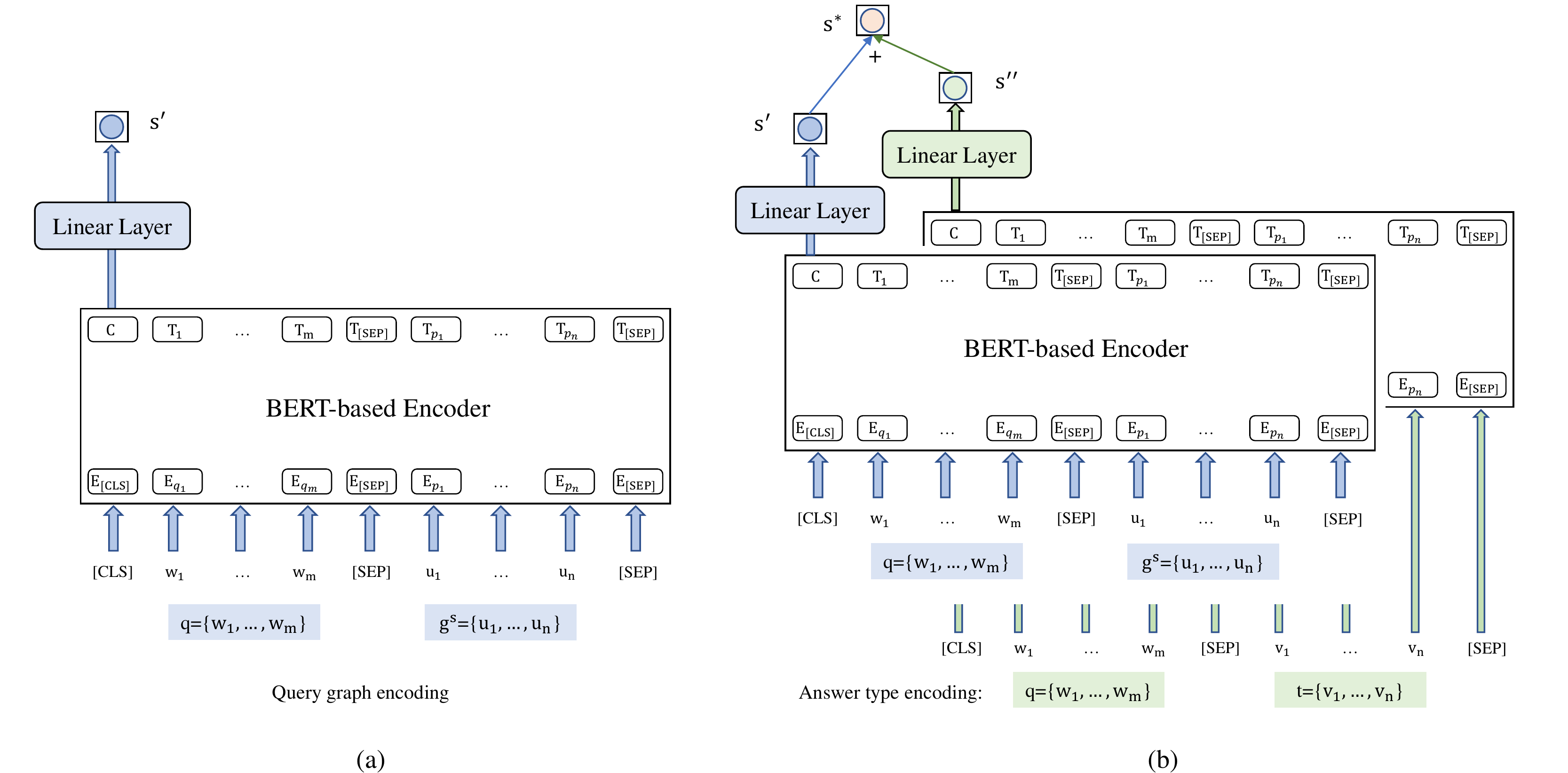}
\caption{(a) The matching framework for query graph ranking. During encoding, the question $q$ and the query graph sequence $g^s$ are considered. (b) The matching framework for query graph reranking. During encoding, the question $q$, the query graph sequence $g^s$ and answer type $t$ are considered.}
\label{fig:rank_frame}
\end{figure*}
After the query graph sequence $g^s$ is obtained, the task of query graph ranking is transformed into a matching task between two sequences: the question and the query graph sequence. We use BERT to encode the question and the query graph sequence. Specifically, for query graph sequence $g^s=\{u_1,u_2,...,u_n\}$, we combine it with the question $q=\{w_1,w_2,...,w_m\}$ to form a sentence pair $q{g}^{s}=\{[CLS],w_1,…,w_m,[SEP],u_1,…,u_n,[SEP]\}$. Then the sentence pair $q{g}^{s}$ is fed into BERT for encoding. The encoding framework is shown in Figure~\ref{fig:rank_frame}(a). We use the output of the $[CLS]$ node in BERT as the semantic representation of the question and the query graph sequence, denoted as $\bm{f}$.
\begin{equation}
\bm{f}=BERT(qg^s).
\end{equation}
The semantic representation vector $\bm{f}$ is mapped to a specific score through a linear layer.
\begin{equation}
s^{'}=\bm{f}W,
\end{equation}
where $s^{'}$ is the similarity score between the question and the query graph, and $W$ is a weight matrix. We calculate the scores of all the candidates in this way and obtain the top-n results.

During training, we sample data according to a fixed positive and negative ratio, as do in Luo et al. (2018)~\cite{luo2018knowledge}. That is, a positive query graph and $N$ negative query graphs form group examples.
In the optimization process, we use the cross-entropy loss function to optimize the score $s^{'}$ of each query graph, and the optimization objective is:
\begin{equation}
s=\frac{1}{1+e^{-s^{'}}},
\end{equation}
\begin{equation}\label{equ:ylabel}
L=-(y\log(s)+(1-y)\log(1-s)),
\end{equation}
where $y$ is the query graph label (1 represents a positive example and 0 represents a negative example).

\subsubsection{Query Graph Reranking}
Obviously, we can rank all the candidate query graphs according to the scores given in query graph ranking, and select the one with the highest score as the optimal query graph. However, in the preliminary experiments, we find that there is a significant gap between top-1 performance and the oracle score of top-n candidates. To further improve top-1 performance, we propose a model to rerank the top-n query graphs and obtain the best query graph.

In query graph reranking, we add the answer type feature to our matching model, where the answer type information is from KB~\footnote{In Freebase, each entity has one or more corresponding type. We use the property value of the relation `fb:common.topic.notable\_types' as the type description of the answer.}. As shown in Figure~\ref{fig:rank_frame}(b), we adopt the same way used in query graph ranking to calculate the matching score $s^{'}$ of the question and the query graph sequence. Besides, we utilize another BERT to encode the question and the answer type. Given a question $q=\{w_1,w_2,...,w_m\}$ and answer type sequence $t=\{v_1,v_2,...,v_n\}$ of the query graph, they are spliced into a sequence $qt$, and then fed into BERT for semantic encoding. We also use the output vector of its $[CLS]$ node as the semantic feature of the question and the answer type. 
\begin{equation}
\bm{f_t}=BERT(qt).
\end{equation}
And then, the feature is mapped to a score through a linear layer.
\begin{equation}
s^{''}=\bm{f_t}W_t,
\end{equation}
where $s^{''}$ is the similarity score of the question and the answer type, and $W_t$ is a weight matrix. Two scores, $s^{'}$ and $s^{''}$, are summed together as the final matching score $s^*$,
\begin{equation}
s^*=s^{'} + s^{''},
\end{equation}

During training, we no longer randomly select negative examples to construct training data like query graph ranking. Instead, we use the top-n examples generated by query graph ranking from training set as training data. 
After introducing the answer type feature, the final optimization objective becomes:

\begin{equation}
s=\frac{1}{1+e^{-s^*}},
\end{equation}
\begin{equation}
L=-(y\log(s)+(1-y)\log(1-s)),
\end{equation}
where $y$ is the query graph label, same as in Equation (\ref{equ:ylabel}).

\section{Experiments}
\subsection{Experimental Setup}
\subsubsection{Datasets.}

\begin{table}
\caption{\label{datasets split} The partitions of WebQ and CompQ datasets. }
\centering
\renewcommand\tabcolsep{6.0pt}
\renewcommand\arraystretch{1.0}
\begin{tabular}{lccc}
\hline {Dataset} & {Train} & {Validation} & {Test} \\ \hline
WebQ & 3,023 & 755 & 2,032 \\
CompQ & 1,000 & 300 & 800 \\
\hline
\end{tabular}
\end{table}

We conduct experiments on two widely used datasets: WebQuestions (WebQ)~\footnote{https://nlp.stanford.edu/software/sempre/} and ComplexQuestions (CompQ)~\footnote{https://github.com/JunweiBao/MulCQA/tree/ComplexQuestions}. These two datasets are based on the form of question-answer pairs, and use Freebase~\footnote{https: //developers.google.com/freebase/} as the knowledge base. The WebQ dataset contains 5,810 question-answer pairs, including simple questions (84\%) and complex questions (16\%). The CompQ dataset is full of complex questions, containing a total of 2,100 question-answer pairs. Following the previous studies, both WebQ and CompQ are divided into training set, validation set and test set. The partitions of the datasets are listed in Table~\ref{datasets split}.

\subsubsection{Implementation Details.}
In query graph ranking and query graph reranking, we both use BERT-BASE for BERT model initialization, and use Adam as the optimizer for optimization. We choose the settings of hyper-parameters in the systems according to the performance of the validation set. The learning rate is set to $5\times10^{-5}$, and the maximum training epoch is set to 5. At the end of each epoch, we evaluate the current model on the validation set and obtain the corresponding score. After training, we select the model with the highest score on the validation set as our final model which is further evaluated on the test set. For the evaluation metrics, we report the average F1-score (F1) that has been used in the previous methods\cite{luo2018knowledge}, and also report the average precision (P) and average recall (R).


\subsection{Main Results}

\begin{table}
\caption{\label{rank-rerank} The comparison results of query graph ranking and query graph reranking on the test sets.}
\centering
\renewcommand\tabcolsep{6.0pt}
\renewcommand\arraystretch{1.0}
 \begin{tabular}{lcccccc}
\hline 
\multirow{2}*{{Method}} & \multicolumn{3}{c}{{WebQ(\%)}} & \multicolumn{3}{c}{{CompQ(\%)}}\\ 
\cmidrule(r){2-4} \cmidrule{5-7}
 & {P} & {R} & {F1} & {P} & {R} & {F1}\\
\hline
Ranking (top-1) & 51.41 & 61.64 & 52.44 & 37.12 & 49.40 & 38.35 \\
Reranking & \bf{54.79} & \bf{65.98} & \bf{56.02} & \bf{41.29} & \bf{55.44} & \bf{42.94} \\
\hline
\end{tabular}
\end{table}
Table~\ref{rank-rerank} shows the performance of query graph ranking and query graph reranking on the WebQ and CompQ datasets, where Ranking (top-1) refers to the system that we only use the ranking step to obtain the answer and Reranking refers to the full system proposed in this paper. From the table, we find that compared to the Ranking system, the Reranking system achieves much better performance. Specifically, we obtain an absolute improvement of 3.58\% F1 on the WebQ dataset and 4.59\% F1 on the CompQ, respectively. In addition, the Reranking system also achieves better performance on precision and recall than the Ranking system. These facts indicate that our reranking solution is quite effective. 
\begin{table*}
\caption{\label{previous-research-compare-table} The comparison results with the previous methods on the test sets. }
\centering
\renewcommand\tabcolsep{4.0pt}
\renewcommand\arraystretch{1.0}
\small \begin{tabular}{llcc}
\hline {Category} & {Method} & {WebQ(F1\%)} & {CompQ(F1\%)} \\ \hline
\multirow{4}{*}{Using Query Graph}
    & Yih et al. (2015)~\cite{yih2015semantic} & 52.5 & - \cr
    & Bao et al. (2016)~\cite{bao2016constraint} & 52.4 & 40.9 \cr
    & Hu et al. (2018)~\cite{hu2018state} & 53.6 & - \cr
    & Luo et al. (2018)~\cite{luo2018knowledge} & 52.7 & 42.8 \cr
    & Lan and Jiang (2020)~\cite{lan2020query} & - & \bf{43.3} \cr
\hline
\multirow{4}{*}{Others}
    & Berant et al. (2013)~\cite{berant2013semantic} & 36.4 & - \cr
    & Jain(2016)~\cite{jain2016question} & 55.6 & - \cr
    & Chen et al. (2019)~\cite{chen2019bidirectional} & 51.8 & - \cr
    & Xu et al. (2019)~\cite{xu2019enhancing} & 54.6 & - \cr
\hline
\multirow{2}{*}{Ours}
 & Ranking & 52.4 & 38.4 \\
 & Reranking & \bf{56.0} & 42.9 \\
\hline
\end{tabular}
\end{table*}


Table~\ref{previous-research-compare-table} shows the comparison results of our systems and the previous systems. According to whether to use query graph, the previous systems are divided into two categories: ``Using Query Graph" and ``Others". In the previous ``Using Query Graph" systems, they utilize a single-stage ranking strategy and adopt the predefined query graph features to improve the performance of query graph selection. Compared with them, our method is based on a two-stage strategy and does not rely on any predefined query graph features. 

As a single-stage solution, our Ranking system achieves comparable performance on the WebQ, but performs slightly lower on the CompQ. This shows that complex questions are more challenging and need to be solved further. Among all the systems, our Reranking system achieves the best result on the WebQ and the second best result on the CompQ, which indicates our reranking solution is quite powerful. 


\subsection{Discussion and Analysis}
In this section, we perform further analysis on the results of our systems. We first discuss the feasibility of the reranking strategy. Then we check the effect of using different components. Finally, we study some cases of using answer types.
\begin{figure*}
\centering
\includegraphics[width=10cm]{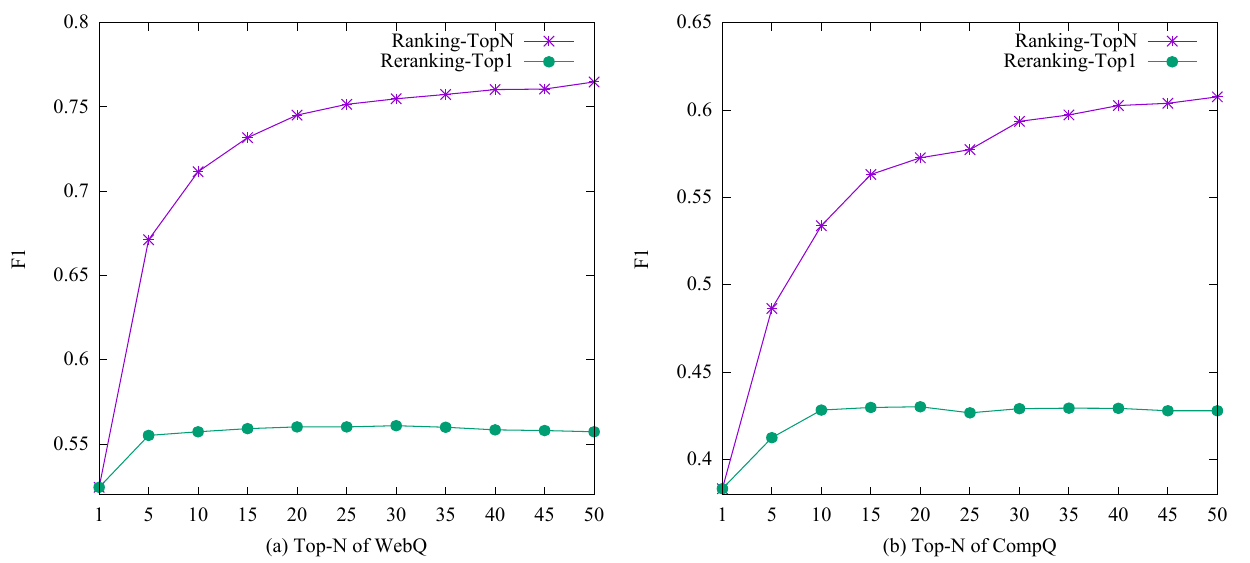}
\caption{The effect of different top-n for reranking, where ``Ranking-TopN" denotes the oracle score of top-n candidates in query graph ranking and ``Reranking-Top1" denotes the top-1 performance of query graph reranking given different top-n results.}
\label{fig:top-n}
\end{figure*}
\subsubsection{Feasibility of Reranking}

We check the oracle score of top-n candidates in query graph ranking, as shown in the  Ranking-TopN curve of Figure~\ref{fig:top-n}. It is obvious that the Ranking-TopN curve at first shows a rapid growth trend, and then tends to be stable when $N>30$. This phenomenon means that most of the correct query graphs are the top candidates in the ranking list, but not always the top-1 one. Therefore, it is possible to use the reranking strategy to improve the performance based on the top-n results.


We further check the performance of reranking with different top-n results given by the step of query graph ranking, as shown in the Reranking-Top1 curve of Figure~\ref{fig:top-n}. From the figure, we find that as top-n increases, the reranking performance at first increases quickly and then remains stable. This means that a reranking method with good performance can be realized under a smaller value of top-n, such as top-10. Thus the reranking solution can efficiently obtain the final answer. We should note that there is room for further improving our system by better top-n ranking model and top-1 reranking model. We leave them as future work.  
\subsubsection{Reasons for reranking promotion}
\begin{table}
\caption{\label{rank-approach} The effect of using different components. }
\centering
\renewcommand\tabcolsep{6.0pt}
\renewcommand\arraystretch{1.0}
\begin{tabular}{lcccccc}
\hline 
\multirow{2}*{Method} & \multicolumn{3}{c}{WebQ(\%)} & \multicolumn{3}{c}{CompQ(\%)}\\ 
\cmidrule(r){2-4} \cmidrule{5-7}
 & P & R & F1 & P & R & F1\\
\hline
Ranking  & 51.41 & 61.64 & 52.44 & 37.12 & 49.40 & 38.35 \\
\hline
Reranking-Base & 53.10 & 63.16 & 53.85 & 38.32 & 50.74 & 39.65 \\
Reranking-Full & 54.79 & 65.98 & 56.02 & 41.29 & 55.44 & 42.94 \\
\hline
\end{tabular}
\end{table}

Compared to the Ranking system, our Reranking system has two changes: we first use a different strategy to construct training data for the matching model, and then add the information of answer type. In order to explore the influence of the above two changes on query graph reranking, we remove the answer type feature from the final system. The results are shown in Table~\ref{rank-approach}, where Reranking-Base refers to the reranking system without the information of answer type and Reranking-Full refers to our final system. From the table, we can find that after removing the answer type, the performance of the system is significantly reduced. But compared to query graph ranking, the Reranking-Base still performs better. This shows that the performance improvement in query graph reranking is not only related to the answer types, but also affected by the reranking strategy itself. 
\subsubsection{Case Study}
Here we perform the case study to know whether the query graph reranking can correct the type error in the query graph ranking with the help of answer types. For the WebQ dataset, we randomly select 50 questions that are wrong in query graph ranking and correct in query graph reranking. The results show that more than half of the examples (52\%) are corrected by choosing the correct answer type instead of the wrong answer type, while others are corrected with the same answer type. Some typical types of corrected examples are listed in Table~\ref{case-table}. 

\begin{table}
\caption{\label{case-table} Some example cases, where italic indicates the relation in the query graph sequence.}
\centering
\renewcommand\tabcolsep{0.0pt}
\renewcommand\arraystretch{1.0}
\begin{tabular}{p{12cm}}

\hline
\textbf{Question1:} where is the Galapagos islands located on a world map?\\
\hline
\textbf{Reranking Select:} Galapagos islands \emph{containedby} Galapagos province, pacific ocean.\\
\textbf{Answer Type:} administrative area, statistical region \\
\textbf{Ranking Select:} Galapagos islands \emph{mentions book} the far side of the world.\\
\textbf{Answer Type:} book \\
\hline
\textbf{Question2:} who is princess leia in star wars?\\
\hline
\textbf{Reranking Select:} princess leia \emph{portrayed in films actor} carrie fisher.\\
\textbf{Answer Type:} person sign, tv actor \\
\textbf{Ranking Select:} star wars \emph{award nominations nominated for} star wars.\\
\textbf{Answer Type:} award nominated work, product category \\
\hline
\end{tabular}
\end{table}
\section{Related Work}
Tracing back to the solutions for knowledge base question answering, there are two mainstream approaches: information retrieval(IR) based methods and semantic parsing(SP) based methods. Among them, IR-based methods~\cite{wang2021retrieval,zhao2019simple} first recognize entities in the question, then search related candidate answers according to entities and returns the most relevant answers. The core of IR-based methods is the matching between questions and answers, which is usually realized by deep learning approaches~\cite{chen2019bidirectional,dong2015question}. For example, Bordes et al. (2014)~\cite{bordes2014question} propose calculate the similarity between the question and the answer by learning low-dimensional embedding of the question and the subgraph of the answer. 
There are also some works that focus on multi-hop reasoning by adopting Key-Value Memory Neural Networks~\cite{xu2019enhancing} or performing a transition-based search strategy~\cite{chen2019uhop}. Besides, knowledge base embeddings are also used to improve multi-hop question answering and achieve success~\cite{saxena2020improving}.

Different from IR-based methods, SP-based approaches pay more attention to the semantic information of questions, trying to parse the question into structured representation~\cite{bao2016constraint,chen2020formal,liang2016neural}. For example, the question can be parsed into $\lambda-DCS$, and then map to the knowledge base to obtain answers~\cite{berant2013semantic}. As for complex questions, Sun et al. (2020)~\cite{sun2020sparqa} design a novel skeleton grammar to express complex questions and improve the ability to parse complex questions. Besides, query graph is also a widely-used formal meaning representation in SP-based systems which is also used in this paper. Yih et al. (2015)~\cite{yih2015semantic} are the pioneer in query graph research for KBQA. Following this line, the encoding method of complex query graph~\cite{luo2018knowledge} and the construction method of multi-hop complex query graph~\cite{lan2020query} are also proposed. In this paper, we propose a two-stage method of query graph selection, which makes the query graph based method more effective.



\section{Conclusions}
In this paper, we propose a simple yet effective method for query graph selection, which can be divided into two stages: query graph ranking and query graph reranking. In query graph ranking, we calculate the similarity between the question and the query graph sequence to obtain the top-n query graph candidates. We further rerank the candidates with the help of the information of answer types. The experimental results show that the method proposed provides the best results on the WebQ dataset and the second best on the CompQ dataset. 

\bibliography{samplepaper}
\bibliographystyle{splncs04}
\end{document}